\documentclass[11pt,a4paper]{article}
\usepackage[hyperref]{acl2021}
\usepackage{times}
\usepackage{latexsym}
\usepackage{amssymb}
\usepackage{graphicx}
\usepackage{multirow,booktabs}

\usepackage{microtype}

\aclfinalcopy 

\title{EXPATS: A Toolkit for Explainable Automated Text Scoring}

\author{Hitoshi Manabe \\
  LegalForce, Inc. \\
  Tokyo, Japan \\
  \texttt{hitoshi.manabe@legalforce.co.jp} \\ \And
  Masato Hagiwara \\
  Octanove Labs \\
  Seattle, WA, USA \\
  \texttt{masato@octanove.com} }

\date{}

\begin{document}
\maketitle
\begin{abstract}
Automated text scoring (ATS) tasks, such as automated essay scoring and readability assessment, are important educational applications of natural language processing. Due to their interpretability of models and predictions, traditional machine learning (ML) algorithms based on handcrafted features are still in wide use for ATS tasks.  Practitioners often need to experiment with a variety of models (including deep and traditional ML ones), features, and training objectives (regression and classification), although modern deep learning frameworks such as PyTorch require deep ML expertise to fully utilize. In this paper, we present EXPATS, an open-source framework to allow its users to develop and experiment with different ATS models quickly by offering flexible components, an easy-to-use configuration system, and the command-line interface. The toolkit also provides seamless integration with the Language Interpretability Tool (LIT) so that one can interpret and visualize models and their predictions. We also describe two case studies where we build ATS models quickly with minimal engineering efforts\footnote{The toolkit is available at \url{https://github.com/octanove/expats}. See \url{https://www.youtube.com/watch?v=B9bmguPS22Q} for video demonstration.}.

\end{abstract}

\section{Introduction}

Automated essay scoring (AES)~\cite{alikaniotis-etal-2016-automatic,taghipour-ng-2016-neural,ke-ng-2019-automated}, text readability/difficulty assessment~\cite{vajjala-meurers-2012-improving,xia-etal-2016-text,vajjala-rama-2018-experiments}, and grammatical acceptability judgement~\cite{heilman-etal-2014-predicting,warstadt-etal-2019-cola} are all important NLP tasks for a wide range of applications including assessment, text simplification, and language education.

All these tasks can be generalized as a task where, given an input text $x$ and an optional context $c$, the model predicts some quality about $x$ as $y = f(x, c)$ where $y$ is a continuous value ($y \in \mathbb{R})$ or a class on an ordinal scale, such as $y \in \mathbb{N}$ or discrete classes (e.g., $y \in \{\mathrm{low}, \mathrm{mid}, \mathrm{high}\}$). For example, in automated essay scoring (AES), $c$ is a prompt (question), $x$ is an input essay, and $y$ is its score. Throughout this paper, we use {\it automated text scoring} (ATS) as an umbrella term to subsume all such tasks.

\begin{figure}[!t]
\begin{center}
\includegraphics[scale=0.38]{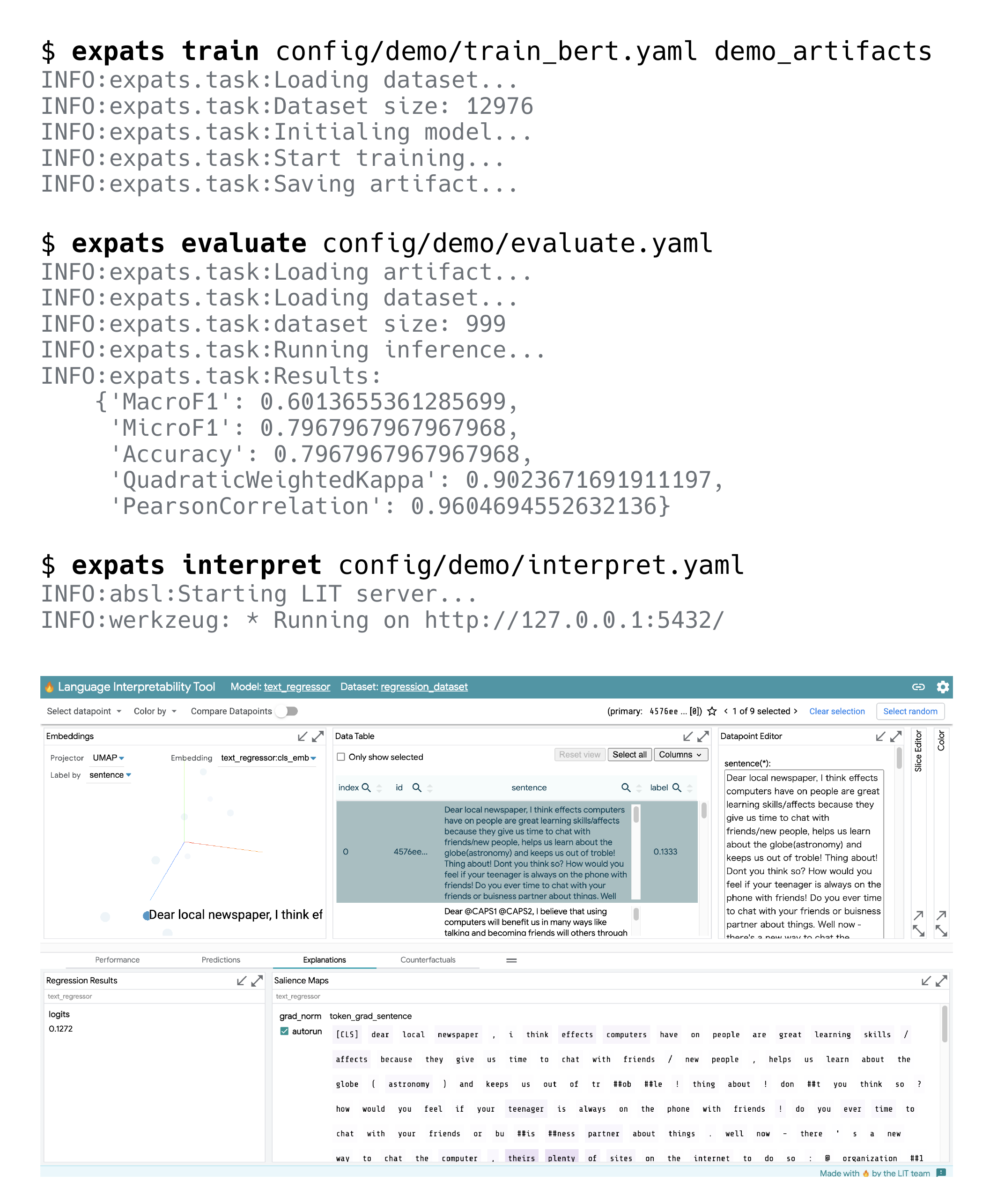} 
\caption{EXPATS command line interface (top) and visualization via LIT~\cite{tenney-etal-2020-language} (bottom)}
\label{fig:overview}
\end{center}
\end{figure}

\begin{figure*}[!t]
\begin{center}
\includegraphics[scale=0.4]{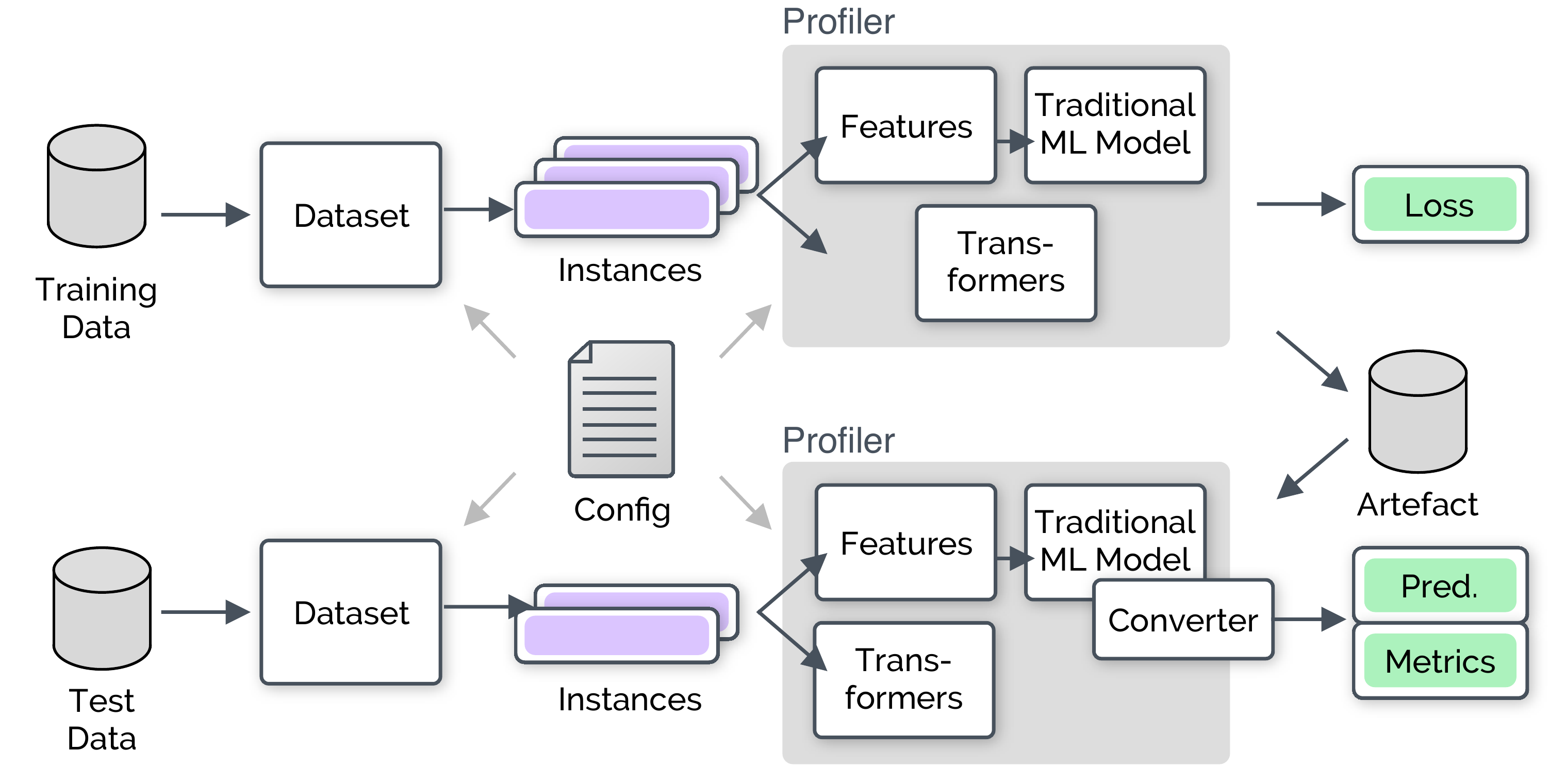} 
\caption{System design and components of EXPATS. See Section 2.1 for more details.}
\label{fig:architecture}
\end{center}
\end{figure*}

Deep neural methods have been used in many NLP tasks including ATS. However, only recently have deep contextualized methods started to be applied to ATS tasks, and their results are still mixed~\cite{nadeem-etal-2019-automated,yang-etal-2020-enhancing,mayfield-black-2020-fine}. In some applications, especially in settings (such as high-stakes AES systems) where fairness and interpretability are important considerations, traditional, machine learning models with handcrafted, well-studied linguistic features are still in use and are preferred\footnote{For a comprehensive summary of feature-based AES methods, we refer interested readers to~\citet{ke-ng-2019-automated}.}. Due to these reasons, practitioners often need to experiment with a wide variety of models, features, and/or training objectives, including regression, (ordinal) classification, and ranking objectives.

Generic deep NLP frameworks such as AllenNLP~\cite{gardner-etal-2018-allennlp} and Transformers~\cite{wolf-etal-2020-transformers} have been proposed and in wide use, but these require deep understanding of machine learning concepts to fully utilize. There are also many application-specific NLP toolkits such as OpenKiwi~\cite{kepler-etal-2019-openkiwi} for quality estimation, OpenNMT~\cite{klein-etal-2017-opennmt} and fairseq \cite{ott-etal-2019-fairseq} for machine translation, and NeuralClassifer~\cite{liu-etal-2019-neuralclassifier} for text classification. However, no general-purpose toolkits exist for automated text scoring, except for EASE\footnote{\url{https://github.com/edx/ease}}, a library for text scoring based on traditional ML, whose scope and usability is quite limited.

In this paper, we present EXPATS (EXPlainable Automated Text Scoring), an open-source toolkit which allows its users to develop and experiment with different automated text scoring models quickly and easily. Its notable features include:

\begin{itemize}
    \item It provides a simple configuration system via human-readable YAML files and an easy-to-use command line interface.
    \item It implements composable and extendable components for both traditional and deep neural models, along with features, training objectives, and metrics commonly used in ATS tasks.
    \item It seamlessly integrates with LIT~\cite{tenney-etal-2020-language} for providing comprehensive interpretation and visualization of the models and their predictions
\end{itemize}

In this paper, we also describe two case studies—one for automated essay scoring with the ASAP-AES dataset and the other for Chinese reading assessment—where we used EXPATS to build competitive ATS models quickly with minimal engineering efforts.

\section{EXPATS}

EXPATS consists of composable and extendable components, a configuration system, and the command line interface (CLI), which all make it easy for practitioners to design and build a wide variety of ATS models and to interpret them. In this section, we'll cover some of the tooklit's technical details.

\subsection{System Design}

Figure~\ref{fig:architecture} shows the overall system design of EXPATS, along with its main components.

\paragraph{Profiler}

The profiler is the core component of EXPATS. By wrapping around an ML model, it produces predictions (e.g., scores for AES) given the input text (e.g., an essay). The design of the profiler is agnostic of the underlying ML frameworks, which means that toolkit users can define and implement their own profiler with a framework of choice, such as PyTorch~\cite{paszke-2019-pytorch} or Scikit-learn~\cite{pedregosa11a}.

EXPATS defines two implementations of the profiler out-of-the-box---feature-based and deep learning profilers. The former produces predictions based on hand-crafted features and traditional ML algorithms implemented in Scikit-learn such as support vector machines (SVMs). The features given to the model are abstracted as features objects, as detailed below. The latter implements neural network-based contexturalizers such as BERT~\cite{devlin-etal-2019-bert} and any other pretrained language models implemented in Transformers~\cite{wolf-etal-2020-transformers}.

Pretrained profilers and model weights are packaged into files called artifacts with their configuration, which can be deserialized and used at the test time to make predictions for the given data. 

\paragraph{Datasets}
A dataset is a collection of instances used for training and evaluating the model, whose design is heavily influenced by how datasets are handled in other ML frameworks such as AllenNLP and PyTorch. Each instance is implemented as a Python dataclass and groups the text input, the label, along with any other extra fields required by the profiler. The toolkit is shipped with dataset readers for common data formats, such as TSV (tab-separated values), as well as a dataset reader for The Automated Student Assessment Prize (or ASAP, a standard corpus for AES\footnote{\url{https://www.kaggle.com/c/asap-aes}}).

\paragraph{Features}
Choice of appropriate features is an important factor for ATS tasks. For example, \citet{xia-etal-2016-text} discuss various types of features for readability assessment extracted from raw text or syntactic trees. EXPATS abstracts features as functions that extracts some useful statistics from the input text and passes them to the profiler. These feature extractors are defined on the top of the analysis results from spaCy\footnote{\url{https://spacy.io/}}, a widely used toolkit for language analysis. 

EXPATS implements a set of basic features by default, as shown below. It is trivial for users to implement their own features by inheriting from the feature class.

\begin{itemize}
    \item Total number of tokens
    \item Average length of tokens (in characters)
    \item Document embeddings
    \item Unigram likelihood (provided by an external dictionary)
\end{itemize}

\paragraph{Objectives}
Different automated text scoring tasks use different training objectives, such as regression~\cite{taghipour-ng-2016-neural}, classification~\cite{vajjala-rama-2018-experiments}, and ranking~\cite{yang-etal-2020-enhancing}, and model developers often need to experiment with more than one. With EXPATS, users can switch between regression (e.g., mean squared errors) and classification objectives (e.g., cross entropy) easily. Their predictions are converted to one another for easier evaluation, as we describe below.

\paragraph{Metrics}
The predictive performance of trained models is usually evaluated using many quantitative measures. EXPATS supports a variety of metrics (abstracted by a Metrics class) widely used in classification or regression settings for ATS, including:
\begin{itemize}
    \item Classification accuracy
    \item Precision, recall, and $F_1$ measure (micro and micro averaged)
    \item Pearson’s correlation coefficient
    \item Quadratic weighted kappa (QWK)
\end{itemize}

A variety of evaluation metrics and training objectives are used for text scoring tasks, including regression-based ones such as correlation coefficients and classification-based ones such as accuracy and quadratic weighted Kappa, or QWK. For example, \citet{taghipour-ng-2016-neural} trained the model with a mean squared error (MSE) loss and evaluated with QWK. \citet{alikaniotis-etal-2016-automatic} also used MSE but evaluated with Pearson's and Spearman's correlation coefficients. 

EXPATS provides an abstract component (``Converter'' in Figure~\ref{fig:architecture}) in charge of converting regression-based continuous predictions into classification-based ordinal labels (for example, by binning continuous values into discrete labels) and vice versa, enabling it to evaluate the model with the same set of evaluation metrics regardless of the training scheme (regression or classification). 

\subsection{Command Line Interface}

EXPATS is equipped with a command line interface (CLI) for running various jobs in an experiment workflow so that its users can develop and experiment with ATS models quickly and easily. The CLI of EXPATS consists of the following four sub-commands:
\begin{itemize}
    \item {\tt\bf train}: trains the model based on the configuration file and saves the result as artifacts
    \item {\tt\bf evaluate}: runs evaluation of a pretrained model with specified evaluation metrics and conversion (e.g., regression to discrete classes)
    \item {\tt\bf predict}: make predictions for given input with a pretrained model
    \item {\tt\bf interpret}: runs the LIT server so that users can visualize and interpret model predictions
\end{itemize}

\subsection{Configuration System}

Users often need to experiment with a wide range of models with different model architectures and hyperparameters, and optimal settings differ vastly from tasks to tasks. In EXPATS, users configure these settings by writing human-readable YAML files, which enables them to run a number of experiments easily without writing Python code. It also encourages model reproducibility and easy tracking of experiments.

The following is an example of a configuration file used for training a transformer-based regression model for the ASAP-AES dataset:

{\small 
\begin{verbatim}
task: regression

profiler:
  type: TransformerRegressor
  params:
    trainer:
      gpus: 1
      max_epochs: 30
    network:
      output_normalized: true
      pretrained_model_name_or_path:
        bert-base-uncased
      lr: 4e-5
    data_loader:
      batch_size: 8

dataset:
  type: asap-aes
  params:
    path: 
      /path/to/training_tsv_file
\end{verbatim}}

There are three main sections in EXPATS configuration files. 
The {\tt task} section specifies the task setting to use (e.g., classification or regression). The {\tt profiler} section defines the type of model or hyperparameters  used to train the model, such as the number of training epochs and the learning rate. Finally, the {\tt dataset} section defines the type of dataset and its parameters to load the data. The {\tt params} sections are interpreted and passed to corresponding Python object constructors.

\subsection{Visualization}

Traditional, feature-based methods such as linear regressions and decision trees are, almost by definition, interpretable, making it easier for developers to see which elements contributes to the model predictions and how. On the other hand, deep neural network-based methods, which are inherently black boxes, have been gaining popularity for automated text scoring tasks. Interpretability is an important factor for text scoring, especially for high-stakes settings such as AES for admission tests and for situations where students wish to receive feedback from the system. To make models and their predictions more interpretable, EXPATS support integration with Language Interpretability Tool (LIT; \citealt{tenney-etal-2020-language}) by default. LIT offers a web-based graphical interface where users can visualize, e.g., saliency maps for tokens that contributed to the prediction. The models built with EXPATS are automatically connected with LIT abstractions and can be visualized and interpreted on a web browser. Figure~\ref{fig:overview} shows a screenshot of visualization via LIT. 

\section{Case Studies}

In the remainder of this paper, we'll describe two case studies where we used EXPATS to build ATS models. Since EXPATS is designed so that practitioners who are not necessarily familiar with NLP research or software engineering can also use, all users need to prepare is the datasets for training and validating the model on, and configuration files. After preparing a YAML file shown in the previous section, one can train a model by running the {\tt expats train} command. The resulting model is stored in an artifact package along with its configuration, and one can evaluate, make prediction from, and/or interpret the model by running {\tt evaluate}, {\tt predict}, and {\tt interpret} commands, respectively.

\subsection{ASAP-AES Scorer}

\begin{table*}[t]
\begin{center}
  \begin{tabular}{lrrrrrrrr} \toprule
    \multicolumn{1}{l}{Model} & 
    \multicolumn{1}{c}{1} & 
    \multicolumn{1}{c}{2} & 
    \multicolumn{1}{c}{3} & 
    \multicolumn{1}{c}{4} & 
    \multicolumn{1}{c}{5} &
    \multicolumn{1}{c}{6} & 
    \multicolumn{1}{c}{7} & 
    \multicolumn{1}{c}{8} \\ \midrule

    Random forest & 0.552 & 0.608 & 0.723 & 0.665 & 0.799 & 0.614 & 0.242 & 0.303 \\
   	BERT & 0.523 & 0.598 & 0.680 & 0.787 & 0.805 & 0.790 & 0.432 & 0.516 \\ \bottomrule

\end{tabular}
\end{center}
\caption{\label{tab:asap-aes}  Evaluation results of QWK scores on test set for ASAP dataset }
\end{table*}

In this first case study, we build an AES system based on ASAP-AES, a de-facto standard dataset published at Kaggle\footnote{\url{https://www.kaggle.com/c/asap-aes}}. We build two regression models based on a feature-based algorithm and a deep learning-based algorithm.

\begin{itemize}
   \item {\bf Random forest}: This model is based on random forest regressor with hand-crafted features. The number of estimators and the maximum tree depth are 100 and 5, respectively. We extracted three linguistic features from a tokenized text: number of tokens, average token length, and average unigram likelihood.  The unigram likelihood of each token is computed from unigram counts obtained from the Tatoeba dataset\footnote{\url{https://tatoeba.org/}}.
   \item {\bf BERT}: This model is based on the pretrained BERT~\cite{devlin-etal-2019-bert} model and fine-tuned on the training corpus for automated essay scoring without any hand-crafted features. We tokenized input texts and prepended with the special token {\tt [CLS]}. The hidden representation aligned at the position of {\tt [CLS]} token is projected into one-dimensional scalar with a linear layer. We used the mean squared error as the objective function.
\end{itemize}

We used 80\% of the published ASAP-AES dataset (which contains 12,978 essays in total) as the training set and 20\% as the test set to evaluate all models. One model is trained for each prompt (question). We followed \cite{taghipour-ng-2016-neural} for all other experimental settings. We describe the results of experiments in Table~\ref{tab:asap-aes}. Overall, the BERT-based method outperforms Random Forest regressor with hand-crafted features.

\begin{figure}[!t]
\begin{center}
\includegraphics[scale=0.28]{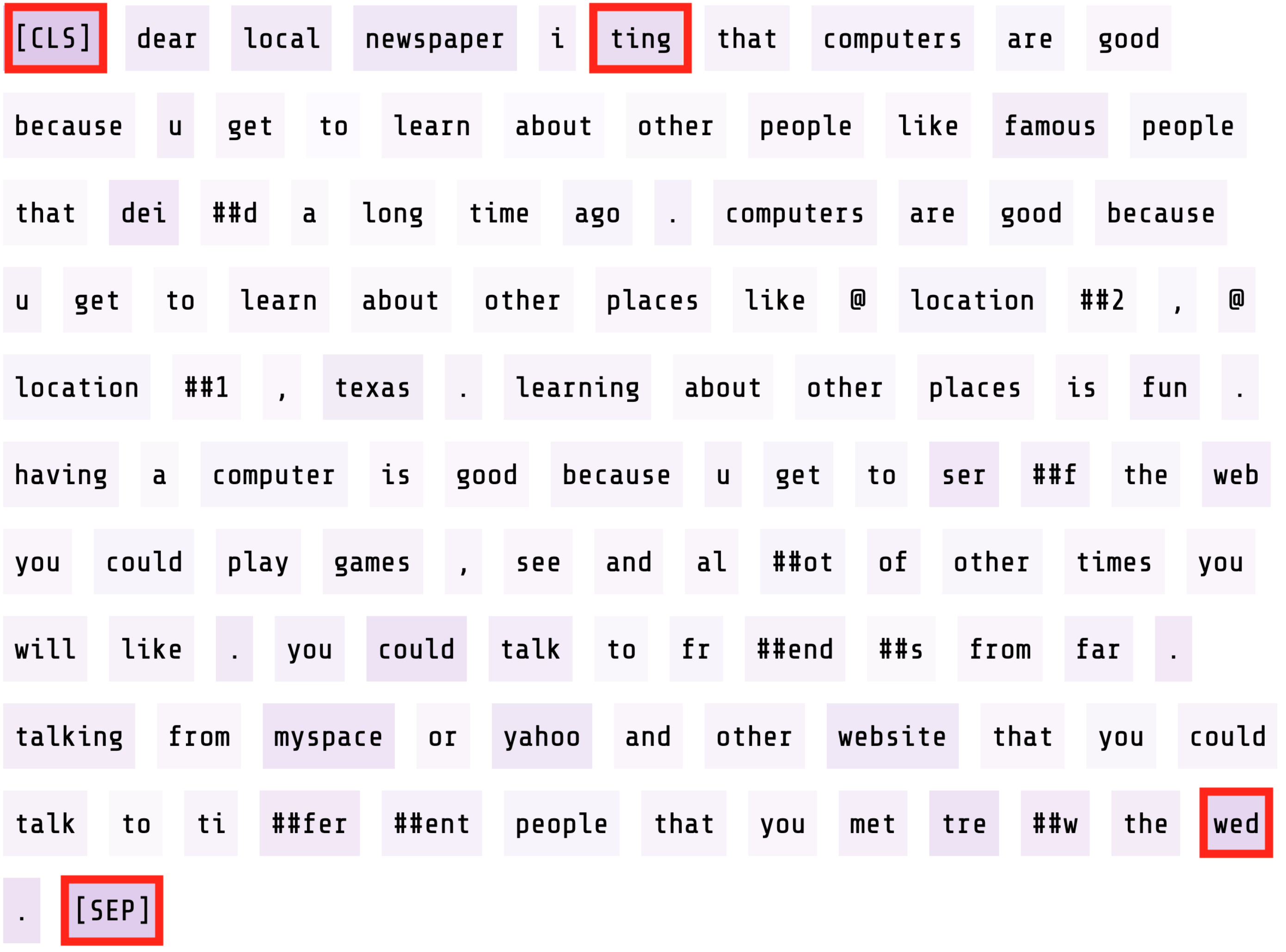} 
\caption{Gradient-based saliency map visualized via LIT. Typos (e.g., {\it ting}) receive higher importance in this low-score essay.}
\label{fig:saliency}
\end{center}
\end{figure}

Next, we qualitatively analyzed the behavior of the trained BERT-based model on the test set using the LIT integration. We inspected saliency maps based on the input gradients, which highlight the importance of input tokens with respect to the model predictions.

In a case where the model predicted lower scores, we find that the tokens containing typos (e.g., {\it ting} in Figure~\ref{fig:saliency}) show more importance than others. On the other hand, infrequent tokens (e.g., {\it predators}) tend to be highly important for the essays with higher scores. This demonstrates that the BERT-based model is making reasonable decisions and the EXPATS-LIT integration is effective for inspecting such relationship.

\subsection{Chinese Readability Assessment}

Readability assessment in other languages has not been explored as much compared to English~\cite{chen-etal-2013-assessing}. In this case study, we will show a case study where we build Chinese readability assessment models based on traditional ML and neural networks (BERT), where the ML models classify a given Chinese passage into six different HSK levels (1-6), which roughly correspond to six levels of CEFR (the Common European Framework of Reference)~\cite{council2001}. As you'll see below, one can easily build models in different languages using EXPATS with minimal modification.

The Chinese dataset for readability assessment we used consists of level-balanced reading passages taken from HSK (a standard Chinese proficiency test) sample questions\footnote{\url{http://www.chinesetest.cn/godownload.do}}. The toolkit supports tab-delimited format of {\tt level[tab]text} out of the box, which means that all we needed to do was prepare their datasets in the same format. 

We compared a random forest classifier as well as a BERT-based classifier. All the other experimental settings are essentially the same as the previous case study. There were only three changes we needed to make to support the Chinese readability assessment:
\begin{itemize}
\item {\bf Tokenizer}: we replaced the spaCy tokenizer model from English to the one for Chinese ({\tt zh\_core\_web\_sm})
\item {\bf Unigram likelihood}: we used a unigram likelihood table computed from a dataset of sentences (approximately 2 million characters) sampled from Wikipedia and Tatoeba.
\item {\bf Contextualizer}: we switched from an English pretrained BERT model to a Chinese one ({\tt bert-base-chinese})
\end{itemize}

\begin{table}[t]
\begin{center}
  \begin{tabular}{lrrrr} \toprule
    \multicolumn{1}{l}{Model} & 
    \multicolumn{1}{c}{Acc.} & 
    \multicolumn{1}{c}{$F_1$} & 
    \multicolumn{1}{c}{QWK} & 
    \multicolumn{1}{c}{Corr.} \\ \midrule

    Random forest & 0.398 & 0.251 & 0.463 & 0.482 \\
   	BERT & 0.584 & 0.468 & 0.755 & 0.762 \\ \bottomrule

\end{tabular}
\end{center}
\caption{\label{tab:chinese} Chinese readability assessment results}
\end{table}

We ran some light hyperparameter tuning on the validation set, and evaluated the performance on the test portion of the dataset. Table~\ref{tab:chinese} shows the experimental results. We again see that the BERT-based neural model achieves better readability assessment performance compared to the random forest, although adding and improving the features used for the traditional ML algorithm will certainly improve its performance. This case study demonstrates that, despite that fact that we dealt with another task in a very different language, we were able to quickly build new models thanks to the EXPATS toolkit's flexibility.

\section{Conclusion}
We presented an open-source framework called EXPATS for automated text scoring (ATS) tasks. EXPATS allows practitioners to develop and experiment with different ATS models quickly and easily, by offering easy-to-use components, the configuration system, and the command-line interface, as well as the integration with LIT for model interpretability and visualization.

We are planning to cover more features and methods (including non-BERT neural networks) with EXPATS as future work. In addition, giving feedback is an important aspect for automated text scoring (see \cite{beigman-klebanov-madnani-2020-automated} for a recent review) and providing more comprehensive visualization not only for model developers but also for language learners is an important venue for future research.

\section*{Broader Impact}

As with other machine learning fields and tasks, fairness and algorithmic biases are an important consideration for ATS tasks and have been discussed intensively in the literature~\cite{beigman-klebanov-madnani-2020-automated}. A scoring system is said to be fair if the score differences are due only to the differences in the constructs (the skills/abilities the system is intended to measure), not due to other indirect factors such as genders or native languages. Common analyses methods for fairness include mean score differences and the model performance for different subgroups~\cite{loukina-etal-2019-many}, and some open-source toolkits exist for assisting fairness-related analyses~\cite{madnani-loukina-2016-rsmtool}. 

The design of EXPATS can contribute to validation and fairness analysis of ATS systems. Its framework (components, the configuration system, and the CLI) enables quick experimentation and validation of various settings, which potentially helps find feature/model biases. It is also straightforward to implement such fairness analysis techniques either directly or via LIT integration. Finally, the EXPATS-LIT integration offers ways for visualizing and identifying sources of potential algorithmic biases. Although little attention has been paid to explainable neural methods for ATS tasks~\cite{kumar-boulanger-2020}, EXPATS can open up a new, important line of research on this front.

\newpage

\bibliographystyle{acl_natbib}
\bibliography{anthology,acl2021}


\end{document}